# What Is Missing in Surgical Risk Stratification and Outcome Prediction: A Scoping Review of End-to-End Machine Learning Approaches

Yizhi Dong, Yuhe Ke, Hairil Rizal Abdullah, Yucheng Xing, Kevan Kai Bing Teo, Ling Huang and Mengling Feng, *Member, IEEE*

***Abstract*—Postoperative adverse events, including mortality and morbidity, remain a major global burden, many of which are preventable with early identification of high-risk patients and targeted perioperative care. Accurate and individualized risk stratification is therefore essential. With the growing availability of large-scale electronic health records (EHRs), machine learning (ML) offers a data-driven approach to model complex clinical patterns and predict surgical outcomes. However, existing studies vary widely in design, and methodological practices across the ML pipeline remain fragmented. This scoping review characterizes and appraises ML pipelines applied to EHR data for surgical risk stratification and outcome prediction. We reviewed 190 studies spanning diverse clinical contexts and covering the full ML workflow, including data preprocessing, algorithm selection, model evaluation, and explainability. Most studies used single-center, private datasets with limited data modalities, and the scarcity of open-access surgical datasets further constrains reproducibility and generalizability. Reporting on key preprocessing steps, including missing data handling, feature selection, and class imbalance, was inconsistent and frequently incomplete. Conventional ML models and simple neural networks predominated, while deep learning and multimodal approaches remained uncommon. Benchmark datasets and standardized evaluation protocols were largely absent, hindering cross-study comparisons. Only about one-third of studies incorporated explainability methods, reflecting growing but inconsistent adoption. This review identifies methodological gaps that currently limit progress toward clinically robust postoperative ML tools. As the first comprehensive methodological review in this area, it provides a structured reference to support more rigorous, reproducible, and clinically meaningful ML development for perioperative care.**

## I. Introduction

EACH year, more than 300 million patients undergo surgical procedures worldwide [1]. Postoperative mortality and morbidity together represent a substantial global burden. In 2019, an estimated 4.2 million people died within 30 days of surgery, accounting for 7.7% of all global deaths [2]. Morbidity is equally concerning, with postoperative complications affecting approximately 14.4% of patients [3]. Evidence shows that current risk stratification is suboptimal. Large cohorts have reported that a large proportion of patients who die after surgery are never admitted to critical care [4], [5], and that many ICU admissions are unplanned, reflecting delayed recognition of high-risk patients [6]. Yet many of these adverse outcomes are preventable through earlier identification of high-risk patients, appropriate allocation of perioperative resources, and timely interventions [4], [7]. These findings underscore the urgent clinical need for accurate, individualized risk stratification. Such tools inform surgical decision-making, support preoperative counselling and patient selection, and guide perioperative management to improve outcomes.

Existing surgical risk scores, such as the American College of Surgeons National Surgical Quality Improvement Program (ACS NSQIP) Surgical Risk Calculator [8] and the Revised Cardiac Risk Index (RCRI) [9], have served as valuable tools for

This work is particially supported by the AI for Public Health Program at Saw Swee Hock School of Public Health, National University of Singapore and by A*STAR, CISCO Systems (USA) Pte. Ltd and National University of Singapore under its Cisco-NUS Accelerated Digital Economy Corporate Laboratory (Award I21001E0002). *(Corresponding author: Ling Huang).*

Yizhi Dong is with the Saw Swee Hock School of Public Health, National University of Singapore, Singapore 117549 (e-mail: yizhi@nus.edu.sg).

Yuhe Ke is with the Data Science and Artificial Intelligence Lab, Singapore General Hospital, Singapore 169608, and also with the Duke-NUS Medical School, Singapore 169857 (e-mail: yuhe.ke36@gmail.com).

Hairil Rizal Abdullah is with the Department of Anaesthesiology, Singapore General Hospital, Singapore 169608, and also with the Data Science and Artificial Intelligence Lab, Singapore General Hospital, Singapore 169608, and also with the Duke-NUS Medical School, Singapore 169857 (e-mail: hairil.rizal.abdullah@singhealth.com.sg).

Yucheng Xing is with the Saw Swee Hock School of Public Health, National University of Singapore, Singapore 117549, and also with the National University of Singapore Guangzhou Research Translation and Innovation Institute, China Singapore Guangzhou Knowledge City, Guangzhou 510700, China (e-mail: xingyucheng@u.nus.edu).

Kevan Kai Bing Teo is with the School of Science, Loughborough University, Loughborough LE11 3TU, UK (e-mail: kevanteo14@gmail.com).

Ling Huang is with the Saw Swee Hock School of Public Health, National University of Singapore, Singapore 117549 and with the Institute of Clinical Sciences, Imperial College London, London SW7 2AZ, UK (e-mail: lhuang@ic.ac.uk).

Mengling Feng is with the Saw Swee Hock School of Public Health, National University of Singapore, Singapore 117549, and also with the Institute of Data Science, National University of Singapore, Singapore 117602 (e-mail: ephfm@nus.edu.sg).

perioperative risk assessment and have been widely adopted in clinical practice. However, the NSQIP risk calculator demonstrates limited performance across diverse populations and surgical types [10], [11], partly due to its reliance on a narrow set of variables. Similarly, the generalizability of the RCRI is constrained, as it was developed from a small cohort of older adults with a low incidence of complications. Moreover, both tools are based on linear or additive assumptions and do not capture complex feature interactions or non-linear relationships. These limitations reduce their predictive power and practical utility in varied clinical contexts.

With the increasing availability of large-scale electronic health records (EHRs), attention has shifted toward data-driven approaches that capture the complexity of surgical risk. Machine learning (ML), a core component of artificial intelligence (AI), has been increasingly applied to healthcare tasks such as diagnosis [12], prognosis [13] and treatment decision support [14]. ML models are particularly well-suited for surgical risk prediction because they can model non-linear relationships, capture high-dimensional interactions, and generate individualized predictions that account for patient heterogeneity and clinical context. Importantly, ML-based risk calculators have demonstrated clear performance gains over traditional linear tools. For example, the ML model in [15] outperformed the NSQIP Surgical Risk Calculator in predicting multiple postoperative complications, underscoring the advantages of moving beyond linear assumptions. Deep learning methods further extend these capabilities by integrating diverse modalities, including structured data such as laboratory results and vital signs, and unstructured data such as clinical notes and imaging. Notably, [16] reported that a deep learning model incorporating dynamic longitudinal clinical measures surpassed physician performance in predicting postoperative acute kidney injury. Beyond accuracy, ML models have shown promise in enhancing clinical decision support (CDS) by enabling timelier care delivery and improving patient outcomes [17]. Recent studies have also integrated ML-based CDS into real-world perioperative workflows, indicating the potential to augment clinical decision-making without compromising care quality [18], [19]. Together, these findings highlight the transformative potential of ML, particularly when combined with EHR data, for surgical risk stratification. At the same time, studies in this domain vary widely in their design, including differences in patient cohorts, surgical procedures, and outcome definitions, as well as in methodological choices around preprocessing, modeling, evaluation, and explainability. This diversity makes it difficult to compare findings across studies or to identify consistent methodological references. A consolidated review of ML implementation is therefore needed to map existing practices and provide a structured reference for researchers and clinicians.

To date, only a limited number of reviews have examined ML for postoperative risk prediction. Yoon and Hassan provided narrative overviews that summarized key AI concepts and ML-based perioperative applications across surgical specialties, offering clinical context and highlighting the breadth of interest in this area [20], [21]. Elfanagely et al. published the first methodological review in 2021, cataloging study contexts such as surgical procedures, sample sizes, input features, and outcomes, as well as methodological aspects including the ML algorithms used and their predictive performance [22]. Bellini et al. expanded this by assessing reporting quality [23], while Loftus et al. emphasized scientific rigor through evaluations of validation practices, reporting of precision metrics for rare outcomes, explainability, and clinical implementation frameworks [24]. While these reviews collectively demonstrate growing recognition of ML's potential for surgical risk prediction, they primarily focus on descriptive summaries of clinical applications, algorithms, and model performance. Systematic assessment that covers the full ML pipeline along with associated methodological choices, including data preprocessing, model development, and post-hoc interpretation, remains limited. This highlights the need for a review that not only summarizes current applications of ML in surgical risk prediction but also evaluates methodological practices across the entire pipeline.

This scoping review aims to provide a comprehensive characterization of current implementations of ML pipelines that leverage EHR data for surgical risk stratification and outcome prediction. In addition, we critically examine methodological practices and highlight persistent limitations that may hinder clinical translation. To achieve this, we synthesize the study contexts, including data sources, surgical procedures, target outcomes, and predictive variables, which together inform the framing of surgical risk prediction problems. We also review data preprocessing strategies, encompassing approaches for handling missing data, selecting informative features, and addressing class imbalance, followed by an assessment of the ML algorithms applied, evaluation metrics reported, and techniques used to enhance model explainability. By integrating these methodological perspectives with applied techniques, this review aims to move beyond descriptive summaries and provide a practical, end-to-end reference for researchers and clinicians working to advance ML-driven surgical risk prediction.

## II. Methods

This scoping review followed the Preferred Reporting Items for Systematic Reviews and Meta-Analyses extension for Scoping Reviews (PRISMA-ScR) guidelines [25]. We searched PubMed, Embase, IEEE Xplore, ScienceDirect, and Google Scholar to identify studies published between 1 January 2013 and 1 June 2023. The search strategy combined two sets of keywords using an AND operator: one related to machine learning and the other to surgical risk stratification or postoperative outcome prediction. The machine learning terms included artificial intelligence, machine learning, deep learning, and neural networks. The surgical prediction terms included surgical risk stratification, surgical outcome prediction, surgical complication prediction, surgical risk prediction, postoperative risk stratification, postoperative outcome prediction, postoperative complication prediction, postoperative risk prediction, perioperative risk stratification, perioperative outcome prediction, perioperative complication prediction, and perioperative risk prediction.

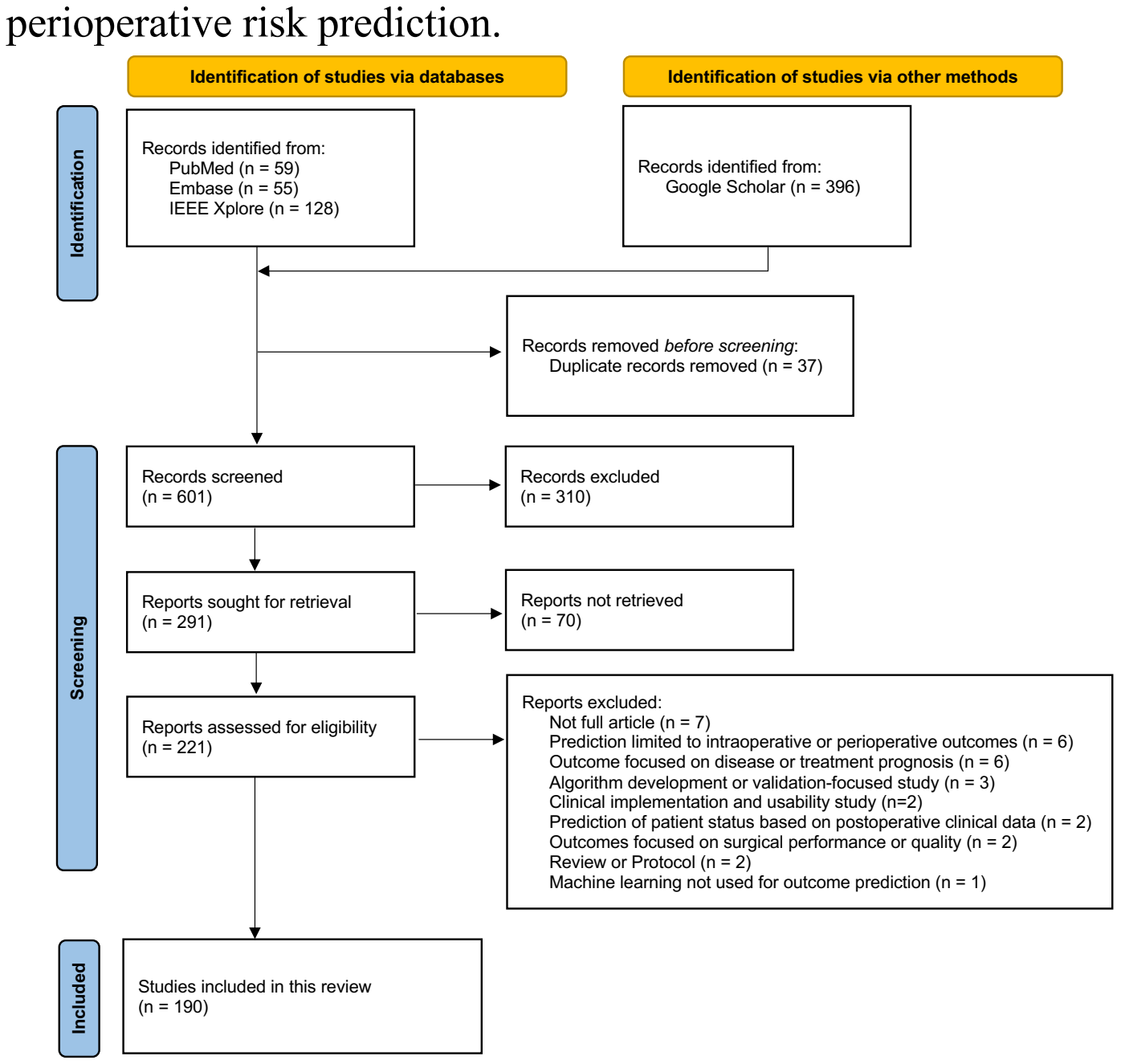

Fig. 1. PRISMA flow diagram of the study selection process.

We included studies that met the following criteria to align with the review's objective: (a) full papers available in English, (b) application of machine learning to predict postoperative risk in surgical settings, and (c) prospective or retrospective study design. All types of machine learning algorithms, surgical procedures, and postoperative risks or outcomes were eligible to provide a comprehensive overview of current practices. The exclusion criteria were defined as follows: (a) studies conducted in outpatient settings, (b) animal studies, (c) papers with fewer than three citations published before 2020, (d) systematic reviews, (e) studies focused strictly on preoperative or intraoperative outcomes, and (f) studies not using data sourced from EHRs or electronic medical records (EMRs).

Title and abstract screening and full-text review were first conducted independently by three authors. Initial screening was performed by title, followed by abstract review for relevance, and full-text assessment for eligibility. A fourth reviewer oversaw the process and resolved any discrepancies. Data extraction was conducted using a standardized form implemented in Excel. For each study, we extracted bibliographic metadata (title, journal, publication year), data source characteristics (dataset, data availability, cohort size), surgical context (procedure type, outcome measures, outcome incidence), input features (feature names, perioperative periods, data modality), preprocessing methods (handling of missing data, feature selection, class balancing), machine learning algorithms, model evaluation (metrics and model performance), and approaches to model explainability. Extracted data were summarized using descriptive statistics, primarily frequency counts. These were calculated across all included studies to identify common patterns and, where applicable, stratified by the most frequently studied surgical procedures. Key distributions and trends were presented using visualizations.

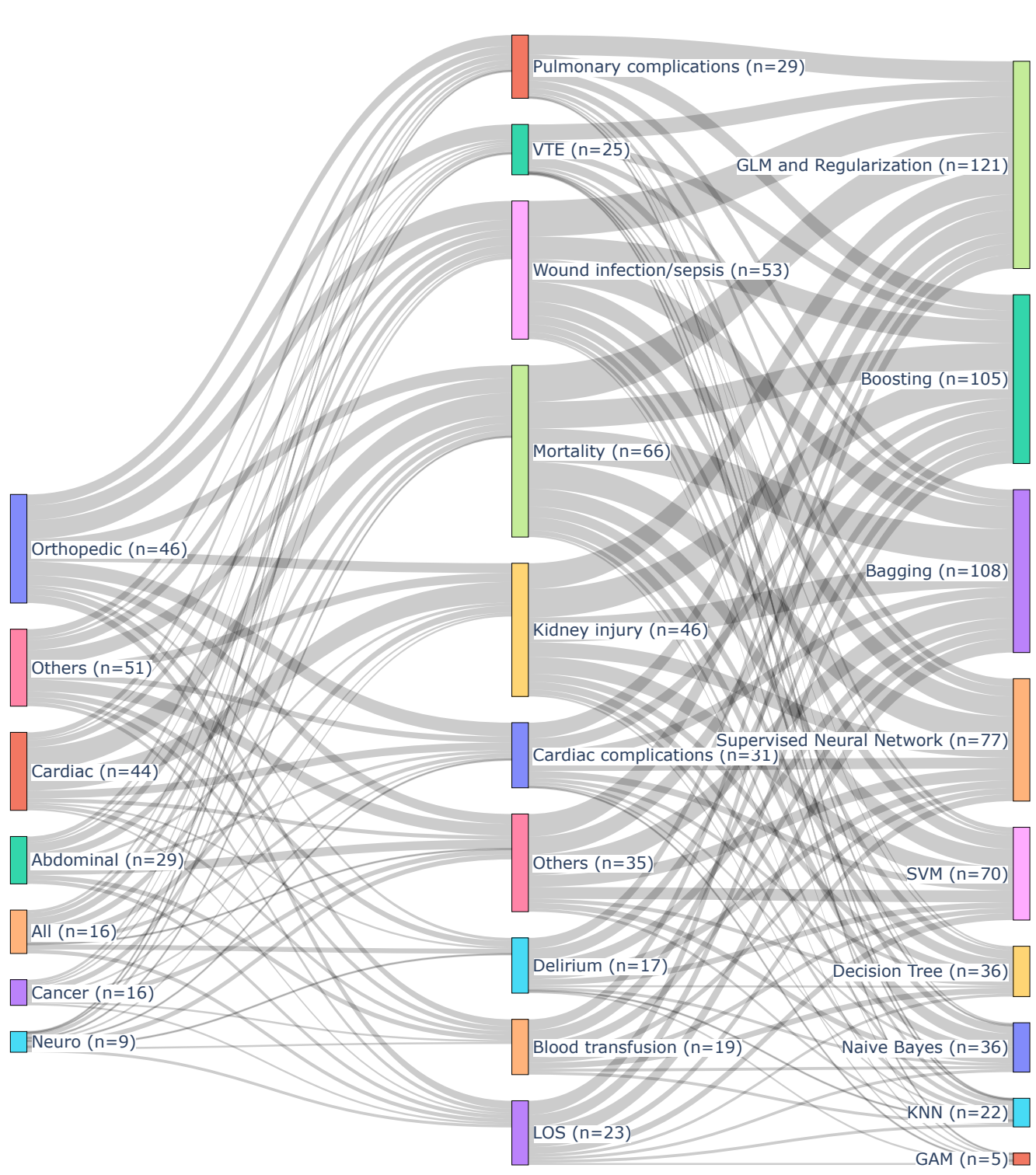

Fig. 2. Sankey diagram illustrating the connections among surgery types, surgical outcomes, and machine learning algorithms.

## III. Results

### A. Study Selection

The study selection process is summarized in Fig. 1 using a PRISMA flow diagram. A total of 638 records were identified through database searches and Google Scholar. After removal of duplicates, records were screened, and reports deemed potentially relevant were sought for retrieval. Following full-text eligibility assessment, 190 studies met the predefined inclusion criteria and were included in the final review.

### B. Data Source Characteristics

All included studies were cohort studies, the majority being retrospective, reflecting the current limitations of applying machine learning in clinical trials. The earliest included study was published in 2015, and the overall publication activity has increased over the past decade (Appendix). Cohort sizes ranged from 32 to approximately 6 million, with a median (Q1, Q3) of 5,757 (897, 43,493). 31 (16.3%) studies included fewer than 500 patients, and 11 (5.8%) had fewer than 200.

Data sources were categorized by scope (single-center vs. multi-center) and accessibility (private vs. open-source). Of the 190 studies, 118 (62.1%) used single-center datasets and 72 (37.9%) used multi-center datasets. Most studies (180, 94.7%) relied on private data, while only 10 (5.3%) used open-source datasets. Notably, none of the included studies used a perioperative-specific public dataset. Instead, only a limited number of publicly available datasets containing perioperative data were identified, including MIMIC [26], [27] and Healthcare Cost and Utilization Project National Inpatient Sample (HCUP NIS) [28].

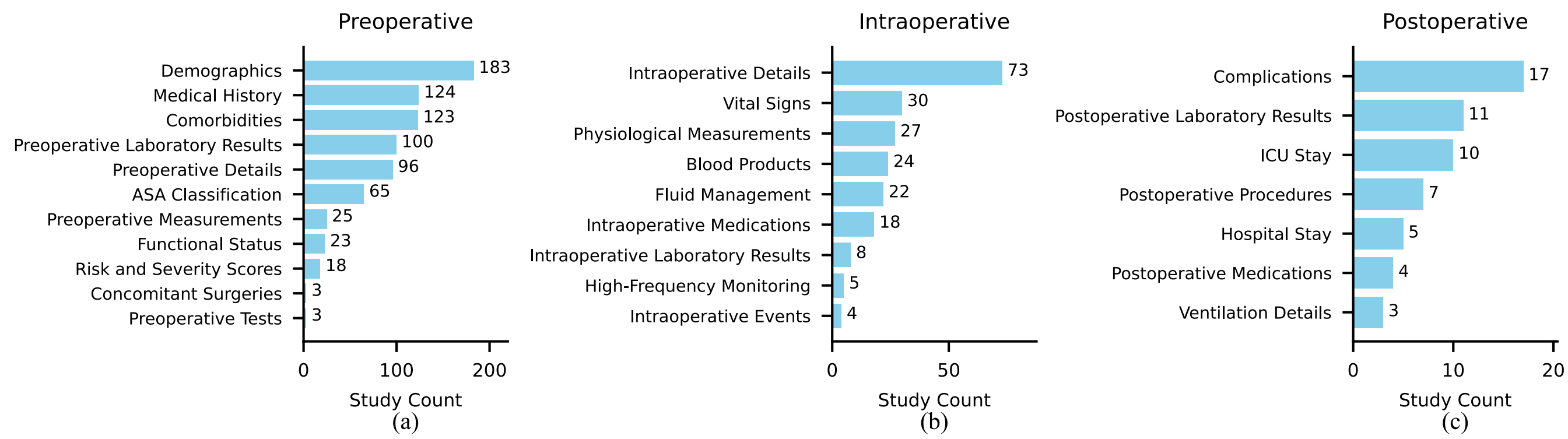

Fig. 3. Summary of input features by clinical categories during the (a) preoperative, (b) intraoperative, and (c) postoperative periods.

## C. Surgical Context

The included studies spanned diverse surgical domains, with some examining multiple procedures. Cardiac and orthopedic surgeries were the most frequently studied procedures (Fig. 2). Studies labeled "All" analyzed surgical outcomes irrespective of procedure type. Common postoperative outcomes included mortality, length of stay, blood transfusion, infection or sepsis, kidney injury, cardiac complications, pulmonary complications, venous thromboembolism, delirium, and stroke. Mortality prediction windows varied from 48 hours to 5 years. The three most frequent study contexts were mortality prediction in cardiac surgery, postoperative kidney injury in cardiac surgery, and infection or sepsis following orthopedic procedures, underscoring the dominance of high-mortality, data-rich specialties.

## D. Input Features

Input variables were categorized by perioperative period of data collection (preoperative, intraoperative, postoperative), data modality, and clinical category. Preoperative features were nearly universal (186, 97.9%), while intraoperative and postoperative data were used in 84 (44.2%) and 34 (17.9%) of studies, respectively. Structured tabular data dominated, reported in 178 studies (93.7%). Smaller subsets of studies used physiological time-series data such as ECG (17, 8.9%), unstructured text such as admission notes (7, 3.7%), or imagingdata such as CT (5, 2.6%). By clinical category (Fig. 3), the most frequent preoperative variables were demographics, comorbidities, medical history, laboratory results, and procedure-related details. The American Society of Anesthesiologists (ASA) physical status classification [29] was included as a preoperative predictor in one-third of studies. Intraoperative variables most often comprised operation time, procedure type, anesthetic and ventilator settings, vital signs, blood loss, ejection fraction, and transfusion records. Postoperative variables primarily included complications, laboratory results, and ICU stay. We also stratified input features by cardiac and orthopedic surgery, the two most common domains (Appendix).

## E. Data Preprocessing and Feature Engineering

### 1) Missing Data Imputation

Missing data are ubiquitous in real-world clinical datasets, where certain variable values may be unmeasured or unrecorded. Incompleteness arises from various factors, including equipment malfunction, patient refusal due to privacy concerns, loss to follow-up, or the absence of clinical indications for specific tests. Since most statistical models and many machine learning algorithms cannot directly handle missing values, imputation strategies are widely applied to replace missing entries with plausible estimates. Among the included studies, 114 (60.0%) explicitly reported the presence of missing data. 22 studies adopted complete-case analysis, excluding records with missing values from downstream analyses, while 93 applied imputation techniques to retain data integrity. Among these 93 studies, imputation strategies were grouped into four categories: univariate statistics-based imputation (45, 48.4%), multiple imputation (29, 31.2%), ML model-based imputation (12, 12.9%), and similarity-based imputation (10, 10.8%).

Univariate statistics-based imputation was the most frequently applied method. In this approach, missing values were replaced with simple summary statistics derived from observed data, such as the mean or median for continuous variables and the mode or a dedicated "missing" category for categorical variables. Multiple imputation provides a more rigorous statistical framework, generating several completed datasets by filling in missing values multiple times to reflect uncertainty in the imputation process. Each dataset is analyzed separately, and results are then pooled according to Rubin's Rules. Multivariate imputation by chained equations (MICE) is the most widely used implementation in the reviewed studies [30], [31], [32]. In MICE, each variable with missing values is iteratively modeled and imputed using predictive models built from the other variables [33]. ML model-based imputation uses machine learning models to iteratively predict missing entries. The most common method identified in the included studies was MissForest [34], [35], which employs random forests as estimators. It performs well on datasets with mixed data types and is effective at capturing non-linear relationships among variables [36]. Similarity-based imputation is another multivariate approach that estimates missing values based on patterns of similar samples. For example, [37] applied the k-nearest neighbors (KNN) algorithm, imputing each missing value by averaging the five nearest neighbors based on Euclidean distance.

2) **Feature Selection**

EHR data are typically high-dimensional and contain substantial noise and redundancy. Feature selection improves model performance and robustness by eliminating irrelevant or redundant variables. Feature selection was reported in 92 studies (48.4%). Among these, the approaches were grouped into four categories: wrapper methods (46, 50.0%), filter methods (30, 32.6%), importance-based methods (25, 27.2%), and embedded methods (23, 25.0%).

Wrapper methods search for optimal feature subsets by iteratively assessing model performance. Common approaches include stepwise selection and recursive feature elimination (RFE). Stepwise procedures add or remove features based on criteria such as the Akaike Information Criterion (AIC) [38] or feature significance based on p-values [39], [40]. RFE ranks features by importance and recursively removes the least informative ones [41]. Random forests were the most frequent base estimator for RFE in surgical risk prediction studies [42], [43], [44].

Filter methods evaluate the relationship between each feature and the outcome independently. For example, [45], [46] applied univariate logistic regression to select statistically significant predictors. [47] employed chi-square tests for categorical variables and F-statistics for continuous variables. Study [48] used Pearson correlation coefficients to rank and select features.

Importance-based feature selection estimates feature contributions from trained models, such as random forests or gradient boosting, and selects the most relevant variables based on predefined thresholds. Unlike filter or embedded methods, this approach does not inherently determine the final subset size, requiring explicit thresholding. Feature importance evaluation is discussed further in the Model Explainability section.

Embedded methods perform feature selection during model training by integrating the selection process with prediction within a single framework. This is typically achieved by introducing regularization penalties that shrink coefficients of less relevant features toward zero. Among the reviewed studies, regularized regression methods were the most widely applied. The Least Absolute Shrinkage and Selection Operator (LASSO) introduces an L1 penalty that drives some coefficients to zero, thereby performing variable selection [49]. LASSO was employed both as a predictive model and as a feature selector in [38], [50], and was applied to derive subsets of predictors for downstream analysis in [51], [52]. Elastic Net, which combines L1 and L2 penalties, was also reported, providing a balance between feature selection and coefficient shrinkage [53].

3) **Data Imbalance**

Class imbalance is a pervasive challenge in surgical outcome prediction. Models trained on imbalanced data tend to favor the majority class, leading to poor representation of minority outcomes. In this review, imbalance was defined as minority class prevalence below 20%. Six studies did not report event proportions and could not be assessed. Among the remaining studies, 131 (71.7%) exhibited class imbalance, including 77 (41.9%) with severe imbalance with minority class prevalence below 5%. Despite this, 75 (56.8%) of studies with imbalance did not implement corrective strategies. Among those that addressed imbalance, 45 (65.2%) applied data-level rebalancing methods, while 11 (15.9%) used algorithm-level adjustments.

Data-level rebalancing strategies included oversampling (39, 86.7%) and undersampling (14, 31.1%), both applied to training sets to achieve a more balanced class distribution. Oversampling augments the minority class either by replicating existing samples or generating synthetic ones. Random oversampling duplicates minority class records, whereas synthetic methods create new data points that better reflect the underlying distribution. In the reviewed studies, 39 papers used synthetic approaches such as the Synthetic Minority Oversampling Technique (SMOTE) [54], Adaptive Synthetic (ADASYN) [55], and Random Over-Sampling Examples (ROSE) [56]. SMOTE was the most widely adopted [57], [58], [59]. It generates synthetic samples with k-nearest neighbors to balance class ratios without creating direct duplications. ADASYN adaptively generates more synthetic samples for harder-to-classify cases [60], [61], and ROSE applies a smoothed bootstrap to create synthetic examples [62]. Undersampling was reported in 19 studies, most often using random undersampling [63], [64], which reduces the number of majority class samples to match minority class frequencies.

Algorithm-level rebalancing assigns greater weight to minority outcomes during training to counter class imbalance. These methods mitigate bias toward the majority class without altering the underlying dataset. One approach is to adjust class weight parameters within machine learning models. For instance, the scale pos weight parameter in eXtreme Gradient Boosting (XGBoost) has been used to up-weight minority samples [30]. Another approach is to modify the loss function. A common example is class-weighted cross-entropy, which adjusts class contributions before aggregating the total loss, and was reported in an image-based surgical prediction study [65].

### *F. Machine Learning Algorithms*

A diverse range of machine learning algorithms has been applied for surgical risk prediction, reflecting differences in data type and task design. Across the included studies, the distribution of machine learning algorithms applied for surgical risk prediction was as follows: supervised learning (148, 77.9%), ensemble learning (133, 70.0%), neural networks (77, 40.5%), and instance-based learning (22, 11.6%). This section summarizes the models most frequently used within these categories (Fig. 2).

1) **Supervised Learning**

Supervised learning was the most widely used framework, where models were trained on labeled datasets to predict surgical outcomes from input features. Reported algorithms included generalized linear models (GLMs) with regularization, support vector machines (SVMs), decision trees, naive Bayes, and generalized additive models (GAMs).

a) GLM and Regularization: GLMs predict outcomes through a linear combination of input features transformed by a link function. Linear regression uses an identity link for continuous outcomes, while logistic regression uses a logit link for binary outcomes. Regularization methods such as LASSO and Ridge add penalty terms to the loss function to prevent overfitting. LASSO applies an L1 penalty that can shrink some coefficients to zero, aiding feature selection, whereas Ridge applies an L2 penalty that reduces coefficient magnitude. Elastic Net combines L1 and L2 penalties, balancing feature selection and coefficient shrinkage.

b) SVM: SVMs are used for both classification and regression tasks. For classification, they identify an optimal hyperplane that separates data into distinct classes, while for regression they estimate a continuous function within a defined error margin. For non-linearly separable data, kernel functions project inputs into a higher-dimensional space where linear separation becomes possible, enabling the modeling of non-linear patterns.

c) Decision Tree: Decision trees are non-parametric algorithms applicable to both classification and regression tasks. They recursively split data based on feature values to form a hierarchy of decision nodes and terminal leaves. Each internal node represents a decision rule, and each leaf node corresponds to a predicted outcome. Optimal splits are determined using criteria such as Gini impurity or information gain.

d) Naive Bayes: Naive Bayes is a classification technique that applies Bayes' theorem to make probabilistic predictions. A Naive Bayes classifier assumes conditional independence among predictive variables. The model combines prior probabilities with observed data to calculate the posterior probability of an instance belonging to each class through the joint probability of the features.

e) GAM: GAMs extend GLMs by incorporating smooth, non-linear functions of predictors, allowing more flexible modeling of complex effects while retaining interpretability.

2) **Ensemble Learning**

Beyond single-model approaches, ensemble methods were widely adopted to enhance predictive stability and accuracy. Ensemble learning is based on the principle that combining predictions from multiple base models yields higher performance and generalizability than relying on a single learner. By integrating diverse model strengths, ensemble frameworks can effectively reduce both bias and variance. Two primary strategies were identified: boosting and bagging.

a) Boosting: Boosting sequentially combines multiple weak learners to form a strong predictive model. Each new learner is trained to correct the errors made by its predecessors, progressively minimizing residuals and improving overall accuracy. Commonly reported algorithms included Adaptive Boosting (AdaBoost) [66], applied in 23 studies, and XGBoost [67], implemented in 64 studies.. AdaBoost iteratively adjusts sample weights to emphasize misclassified instances, whereas XGBoost enhances computational efficiency and model robustness through regularization, parallelization, and automatic handling of missing data.

b) Bagging: Bagging, or bootstrap aggregating, aims to improve model stability and reduce variance in both classification and regression tasks. It generates multiple bootstrapped subsets from the original dataset, trains base models independently on each subset, and aggregates their predictions to enhance robustness and mitigate overfitting. Random forest [68] was the most commonly used bagging algorithm, appearing in 115 studies and employing decision trees as base learners.

3) **Neural Network**

Neural networks use stacked computational layers to learn complex mappings from inputs to outputs. Each layer applies non-linear transformations to intermediate representations, enabling the model to capture dependencies across diverse features. While neural networks can achieve strong predictive performance, their complexity often limits interpretability, leading them to be marked as "black-box" models. Supervised neural networks learn mappings from input features to labeled outputs by minimizing a loss function, applied in 77 studies. Most adopted relatively simple architectures, such as multilayer perceptrons (MLPs), to process structured data. Only a few studies tailored network designs to specific prediction tasks and incorporated diverse data modalities. Five studies modeled physiological time-series data using neural networks. All employed recurrent architectures, including recurrent neural network (RNN) [16], [69] and its variants such as long short-term memory (LSTM) [70], gated recurrent unit (GRU) [71], and GRU with decay (GRU-D) [72], to capture temporal dependencies. One study implemented a dual-pathway model combining CNN and LSTM to extract complementary time-series features [73]. Imaging data were used in three studies. [74] applied DenseNet for CT slice selection and U-Net for tissue segmentation to quantify body composition. [65] used U-Net for surgical site infection detection, while [75] implemented a CNN to classify postoperative complications. Unstructured text was incorporated in only one study. Word embeddings generated by the fastText algorithm were used as input to a CNN with a self-attention module for outcome prediction [76].

4) **Instance-Based Learning**

Instance-based learning is a non-parametric approach that makes predictions by measuring the similarity between new inputs and stored training examples. Unlike other methods, it does not build an explicit model during training. Instead, it memorizes the training data and performs the computations during prediction. K-Nearest Neighbor (KNN): KNN is a classic supervised learning algorithm used for both classification and regression. For classification, it assigns the label most common among the k nearest neighbors. For regression, it predicts the outcome as the average value of the k nearest neighbors.

## G. Model Evaluation

Effective evaluation metrics are crucial for machine learning methods used in surgical risk stratification and outcome prediction, as they directly impact the accuracy and reliability of predictions. These metrics ensure that models can accurately distinguish between different risk levels and predict surgical outcomes with confidence, thereby improving patient safety and enabling personalized treatment plans. Additionally, effective evaluation metrics are essential for meeting regulatory standards and maintaining ethical practices in clinical settings.

Among the reviewed literature, the commonly reported evaluation metrics included accuracy, sensitivity, specificity, precision, recall, F1 score, the area under the receiver operating characteristic curve (ROC-AUC), the area under the precision–recall curve (PR-AUC), positive predictive value (PPV), negative predictive value (NPV), and the Brier score. Accuracy reflects the overall proportion of correct predictions. Sensitivity and specificity quantify the ability to correctly identify true positives and true negatives among all actual positives and negatives, respectively. ROC-AUC, computed as the area under the curve of sensitivity versus 1−specificity across thresholds, summarizes discrimination between positive and negative cases. PPV and NPV describe the probability that a predicted positive or negative is truly positive or negative. Precision (equivalent to PPV) is the proportion of true positives among predicted positives, and recall (equivalent to sensitivity) is the proportion of true positives among all actual positives. The F1 score is the harmonic mean of precision and recall, providing a single balanced measure. PR-AUC is the area under the precision–recall curve emphasizing performance on the positive class, and is therefore particularly informative when the positive class is rare. Across studies, ROC-AUC was the most frequently reported metric (117, 93.7%), followed by sensitivity (86, 45.8%), specificity (80, 42.63%), and accuracy (63, 33.2%). Other metrics were less commonly used, including PPV (44, 23.2%), NPV (35, 18.4%), F1 score (34, 17.9%), PR-AUC (25, 13.2%), precision (24, 12.6%), and recall (16, 8.4%). In addition to evaluating predictive performance, researchers also assess the reliability of predicted probabilities through calibration. This was reported in 39 (20.5%) reviewed studies, most commonly using calibration plots and the Brier score. Calibration plots visualize the agreement between predicted probabilities and observed outcomes. The Brier score quantifies this agreement as the mean squared difference between predicted probabilities and actual outcomes.

## H. Model Explainability

Model explainability is essential in medical contexts, particularly for surgical outcome prediction. It allows clinicians to understand how models arrive at predictions, assess key risk factors, detect potential biases, and evaluate prediction reliability. These capabilities are critical for building trust and ensuring transparency. Among the reviewed studies, 56 (29.5%) reported using explainability methods, most of which focused on feature importance analyses applied to conventional machine learning models. Among these 56 studies, common techniques included Shapley Additive Explanations (SHAP) (26, 46.4%), impurity-based importance (16, 28.6%), coefficient-based importance (10, 17.9%), and perturbation-based importance (3, 5.4%). Three studies (5.4%) applied explainability techniques to deep learning models.

### 1) SHAP

SHAP is the most frequently used method to evaluate feature importance. It applies game theory to assign each feature a contribution value for a specific prediction, making it suitable for complex models such as ensemble methods and neural networks [77]. At the global level, SHAP values quantify how each feature contributes to predictions across the dataset, both in direction and magnitude. Locally, SHAP explains feature impact for individual predictions. A key property is additivity: the sum of SHAP values equals the difference between the model's output and a baseline prediction. For instance, SHAP was applied to a gradient boosting model in [47] to identify influential features affecting outcomes after pancreatectomy.

### 2) Impurity-Based Feature Importance

This method is used in tree-based models to evaluate how much each feature reduces label impurity. In each split, the selected feature maximizes impurity reduction, typically measured by the Gini index. In ensemble models such as random forests, feature importance is computed by aggregating these reductions across all trees. For example, [78] calculated importance scores by summing Gini reductions across trees, while [79] considered the frequency of feature usage in XGBoost splits. However, impurity-based importance reflects only relative contribution and does not indicate directionality.

### 3) Coefficient-Based Feature Importance

Linear models such as logistic regression are inherently interpretable. Feature importance is directly inferred from model coefficients or odds ratios (ORs), which describe the magnitude and direction of association with the outcome. Positive coefficients (OR > 1) indicate a positive association, while negative coefficients (OR < 1) indicate a negative one. However, relying solely on point estimates can be misleading, as wide confidence intervals may indicate low statistical reliability. Confidence intervals and p-values are therefore essential for interpreting the significance of each coefficient. This method is also sensitive to data quality issues such as outliers and missingness, necessitating careful preprocessing

### 4) Perturbation-Based Feature Importance

These methods evaluate how model performance changes when input features are modified. A common example is permutation importance, which randomly shuffles feature values to assess how performance degrades, as applied in [35]. Another approach removes individual features from the model and quantifies the performance drop, as demonstrated in [80]. These methods indicate the model's reliance on specific features but do not reflect the direction of association.

### 5) Explainability Methods for Deep Learning Models

Deep neural networks are often perceived as 'black boxes' due to their multi-layered and non-transparent structure. Explainability methods are therefore critical for building clinical trust, identifying key predictive patterns, and detecting potential biases. However, only a few studies applied such methods to deep learning models. SHEM introduced a self-explaining module alongside a recurrent layer to enable end-to-end explainability through linear approximation [81].

MediMLP applied Gradient-weighted Class Activation Mapping (Grad-CAM) to identify variables most associated with postoperative complications [82]. In [71], the authors used Integrated Gradients to attribute importance both at the feature level and across time steps.

## IV. Discussion

This review work presents the first comprehensive methodological review of ML applications for surgical risk stratification and outcome prediction. In this scoping review, we systematically synthesized and critically evaluated 190 studies leveraging EHR data to support postoperative risk prediction, with a particular focus on end-to-end methodological practices. Publication activity has generally increased over the past decade, reflecting the growing integration of ML into perioperative medicine. However, this expansion has not been matched by corresponding methodological maturity.

### A. Data Source and Surgical Context

Cohort sizes varied substantially, and a notable number of studies relied on relatively small samples, with some including fewer than 200 patients. The use of small datasets raises concerns about overfitting, model instability, and limited generalizability, particularly when the number of features is large relative to the available sample size. Furthermore, most studies were conducted in single centers using privately held datasets, limiting reproducibility and external validation across heterogeneous populations. The scarcity of publicly available perioperative datasets underscores the absence of standardized benchmarks, which impedes systematic performance comparison and reproducibility. Encouraging the creation, sharing, and adoption of large-scale, diverse, and standardized benchmark datasets will be an essential step toward improving transparency and comparability. Recent open-access initiatives such as INSPIRE, VitalDB, and MOVER offer promising starting points for this effort [83], [84], [85].

Although structured tabular EHR data were almost universally used, the integration of unstructured text, physiological time-series, and imaging data remains limited. This underutilization persists despite rapid methodological advances in natural language processing (NLP), time-series analysis, and computer vision (CV), all of which have demonstrated strong capabilities for extracting clinically meaningful information [86], [87], [88]. Given the inherently multimodal nature of perioperative data, the narrow reliance on structured inputs represents a missed opportunity to enhance predictive performance and model robustness. Future research may benefit from greater efforts toward curating and sharing multimodal datasets, which could facilitate the development of models that better capture the complex physiological, temporal, and contextual factors shaping surgical outcomes.

Surgical contexts varied considerably across studies, with orthopedic and cardiac procedures being the most frequently investigated. Among specific prediction targets, mortality and kidney injury in cardiac surgery, as well as wound infection or sepsis in orthopedic surgery, were the most commonly modeled outcomes. This predominance likely reflects both their broad clinical relevance and the pressing demand for early risk identification in these populations. Most studies focused on preoperative data, aiming to enable early detection of high-risk individuals before surgery. Demographic characteristics, medical history, and comorbidities were the most frequently utilized preoperative features. Approximately half of the studies incorporated intraoperative variables, such as surgical details, vital signs, and physiological time-series data. Several studies reported improved predictive performance when intraoperative features were combined with preoperative data, suggesting that intraoperative information captures dynamic physiological changes and procedure-specific events that enhance perioperative risk stratification [73], [89], [90]. In contrast, postoperative data were less frequently utilized, as they are typically relevant for predicting long-term adverse outcomes. When included, postoperative features most often comprised postoperative complications, laboratory results, and ICU stay-related variables.

### B. Data Preprocessing and Feature Engineering

Data preprocessing is a critical yet often underreported stage of the ML development pipeline. Surgical risk prediction encompasses key steps such as handling missing data, selecting informative features, and addressing class imbalance.

Missing data were common across studies; however, almost half did not report whether missingness was present. In approximately one-tenth of the studies, records with missing feature values were excluded from analysis, leading to potential information loss and bias, particularly when the missingness is not random. Among studies that explicitly addressed missingness, univariate statistics–based and multiple imputation were the most frequently applied techniques, indicating that conventional statistical approaches remain predominant. Univariate imputation is computationally efficient but ignores inter-variable dependencies, which can distort feature distributions and attenuate model performance. Multiple imputation (e.g., MICE) provides theoretical rigor and uncertainty quantification but can be computationally infeasible for high-dimensional or large datasets [91]. Machine learning–based imputation methods, particularly MissForest, were adopted in several reviewed studies. These methods capture nonlinear feature dependencies and generally offer greater flexibility than traditional statistical models. In contrast, deep learning–based approaches were not applied in the reviewed surgical prediction studies. While they represent a methodological advance, they are often more computationally intensive and harder to optimize. Still, deep learning–based imputation has been evaluated in broader data settings. Comparative benchmarking studies have reported mixed findings. Some indicate that deep learning methods, such as Generative Adversarial Imputation Networks (GAIN), outperform both statistical and conventional ML-based approaches across a range of data characteristics [91]. Others show that conventional methods like MICE and MissForest remain more robust and effective for small to moderate-sized tabular datasets [92]. Overall, the effectiveness of imputation techniques appears dependent on data context, such as sample size, feature types, missingness patterns, and feature distributions. To strengthen methodological transparency and model reliability, future studies should explicitly report data completeness and imputation procedures. They could also evaluate how imputation choices interact with specific task settings to influence model robustness and predictive performance.

Feature selection was explicitly reported in fewer than half of the reviewed studies, while others selected variables based primarily on clinical expertise or prior evidence. Feature selection enables data-driven exploration of a broader set of variables while eliminating irrelevant or redundant inputs. By reducing input dimensionality, it can enhance model generalizability and computational efficiency,and is particularly beneficial in small-sample or imbalanced datasets where overfitting is a concern. Among studies employing feature selection, wrapper methods were the most prevalent, followed by filter, importance-based, and embedded approaches. Filter methods are computationally efficient and model-agnostic but may overlook multivariate feature interactions. Wrapper methods evaluate feature subsets using model-based training, capturing feature dependencies but often at the cost of computational expense and potential overfitting. Embedded methods perform feature selection during model training, offering an integrated and efficient alternative. However, the selected features are often specific to the algorithm used and may not reflect generalizable variable importance. Importance-based methods provide an interpretable, flexible, and computationally efficient way to rank features across different model types, but they require careful threshold setting to balance the risks of overfitting and excluding informative variables. Some studies combined multiple selection strategies to leverage the complementary strengths of different approaches, forming hybrid methods. In clinical applications, it is also important to balance model-derived relevance with established clinical knowledge, as features with limited importance in data-driven selection may still carry recognized clinical value. Nevertheless, several well-known feature selection algorithms remain underutilized in this field. Filter method such as mutual information [93] quantify the overall dependence between features and outcomes. Wrapper-based methods like the Boruta algorithm [94] test feature relevance by comparing importance values against randomly permuted copies, while evolutionary algorithms such as genetic algorithms [95] iteratively evolve optimal subsets through heuristic search. Future studies could evaluate feature selection strategies based on their predictive utility, robustness, and consistency with clinical understanding, to ensure that selected variables support both accurate modeling and meaningful clinical interpretation.

Class imbalance was a prevalent challenge. Several studies did not report event incidence, and a majority focused on outcomes with low prevalence, including a substantial number targeting events with less than 5% incidence. Despite this, more than half of the studies facing imbalance did not implement any remedial strategy. In such settings, rare but clinically consequential events are underrepresented during model training, potentially biasing predictions and undermining clinical reliability. To mitigate imbalance, both data-level and algorithm-level strategies were adopted. At the data level, oversampling was most common, followed by undersampling. Oversampling increases minority-class representation but may raise overfitting risk through duplicated instances. Synthetic oversampling methods (e.g., SMOTE, ADASYN) help reduce this risk by generating new samples, though they should be applied only within training data to avoid information leakage. Undersampling reduces the majority class to balance the dataset and is computationally efficient but sacrifices potentially informative samples. At the algorithm level, adjusting class weights was the most common approach, assigning greater penalties to minority-class errors during training. While these strategies were variably implemented and reported, consistent documentation of outcome distributions and imbalance-handling methods would improve transparency. In surgical risk prediction, the minority class often corresponds to adverse outcomes that carry clinical importance. Addressing imbalance is essential to ensure the model captures sufficient signal from these events. Beyond widely adopted techniques, ensemble-based approaches such as Balanced Random Forests [96] and RUSBoost [97] remain promising yet underexplored, as they mitigate the dual risks of information loss and overfitting by aggregating models trained on multiple resampled subsets. Future work should also evaluate how imbalance-handling strategies interact with specific surgical outcomes and dataset characteristics to guide method selection.

### C. Machine Learning Algorithms

In this review, machine learning methods for surgical risk prediction were categorized into four broad types: supervised learning, ensemble learning, neural networks, and instance-based learning. Within this framework, we summarize the most frequently used algorithms, ranked by their reported use. The most commonly applied method was the generalized linear model (GLM) with regularization, a supervised learning approach. Its widespread use reflects the method's interpretability and long-standing role in clinical research. GLMs allow direct estimation of feature–outcome associations, which supports clinical reasoning and transparency. However, their reliance on linear assumptions limits their ability to model complex or higher-order interactions. The second and third most common approaches were bagging and boosting, belonging to the ensemble learning category. These approaches typically offer strong predictive performance on structured data by modeling nonlinear relationships, and provide moderate interpretability through feature importance metrics. Among them, XGBoost was widely used for its balance of performance and computational efficiency, and it often outperformed more complex deep learning models in tabular prediction tasks [98]. Neural networks appeared in approximately 40 percent of the studies, all within a supervised learning framework. Most implementations used simple architectures such as multilayer perceptrons, and few studies customized network structures for specific tasks. Deep learning models often faced performance constraints due to small sample sizes and heterogeneous data quality. Their lack of inherent interpretability also limits clinical adoption without additional explanation techniques. SVMs, a type of instance-based learning method, were also widely used. They are effective for small-sample, high-dimensional datasets, but present challenges in large-scale applications due to computational demands and sensitivity to kernel selection. Compared to tree-based models, SVMs also offer lower interpretability. Overall, regression-based and conventional machine learning models still dominate surgical risk prediction, likely due to the tabular nature of most perioperative EHR datasets. While deep learning has traditionally been used for modalities such as imaging and unstructured text, its growing presence in recent studies suggests a gradual shift toward more flexible, data-driven modeling. Recent advances have produced architectures designed for tabular data offering competitive performance, such as transformer-based models [99]. These approaches offer promising advantages, such as reduced need for manual feature engineering, the capacity to model complex interactions, and greater flexibility in integrating high-dimensional or multimodal inputs.

### D. Model Evaluation and Explanation

Evaluation of surgical-event prediction models should consider not only discrimination but also clinical relevance. In our review, the most frequently reported metrics were ROC-AUC, sensitivity, specificity and accuracy. ROC-AUC is frequently used as a primary evaluation metric that quantifies discrimination between positive and negative cases, and it is often supplemented by additional measures to capture performance aspects aligned with clinical priorities. In surgical contexts, low sensitivity implies missed detections of adverse events, whereas low specificity leads to unnecessary interventions. Accuracy, though common and intuitive, was reported in only one-third of studies; it can underrepresent minority-class performance, which can be misleading when rare positive cases are the focus. PPV and NPV are directly interpretable in clinical decision-making, as they reflect the probability that a predicted positive or negative result is correct. When rare positive cases are the focus, metrics on the precision–recall scale are more clinically informative as they both quantify the proportion of true positives. In surgical risk prediction, precision reflects the credibility of a high-risk prediction, whereas recall reflects the effectiveness of high-risk event capture. The F1 score summarizes the trade-off between the two. Correspondingly, PR-AUC is often preferred over ROC-AUC in highly imbalanced settings, as it emphasizes the positive class and provides a more robust assessment in such contexts. Beyond discrimination metrics, calibration assesses the agreement between predicted probabilities and observed outcomes. Well-calibrated models are essential for clinical risk prediction, as inaccurate probability estimates may mislead decision-making, potentially resulting in inappropriate preoperative planning, suboptimal resource allocation, and even harmful clinical decisions [100], [101]. Despite this, calibration was assessed in only about one-fifth of the reviewed studies. Future work on ML models to support surgical risk decision-making should report calibration in addition to discrimination metrics and apply recalibration methods (e.g., isotonic regression) to ensure clinically reliable probability estimates. Furthermore, our findings highlight the need for unified benchmarks. Although all studies reported model performance and some compared multiple algorithms, heterogeneity in prediction outcomes, datasets, and evaluation metrics limits meaningful comparison across studies. The absence of benchmarking datasets and standardized evaluation procedures further constrains the ability to discern model performance differences and track progress in the state of the art.

Explainability is a key requirement for clinical adoption, as it allows practitioners to understand which factors most strongly influence the predicted risk, and identify discriminatory bias or inappropriate reasoning in model decisions [102]. Yet fewer than one-third of the reviewed studies discussed explainability, suggesting that interpretability is gaining attention but remains secondary to model performance in current research. Traditional machine learning methods, such as logistic regression and decision trees, inherently offer interpretable outputs through such as coefficients or rule-based structures. These methods enable clinicians to directly link variables such as age, comorbidities, or laboratory values with predicted surgical outcomes, thereby validating whether the model aligns with established clinical knowledge. Even for more complex models like ensemble models, post-hoc explainability techniques, such as permutation importance or SHAP values, can quantify the relative contribution of input features, highlighting key risk drivers in a transparent and clinically meaningful manner. This form of explainability supports not only clinical trust but also facilitates bias detection, subgroup analysis, and model refinement, ensuring that predictions are both actionable and reliable. In contrast, deep learning methods offer strong predictive power by capturing complex nonlinear relationships and temporal dynamics, but their interpretability is less straightforward. These models are often perceived as “black boxes”, which poses challenges for clinical trust and adoption. Efforts to improve interpretability in deep learning have focused on mechanisms such as attention [103], saliency mapping [104], and surrogate explanations (e.g., LIME [105], SHAP [77]), which aim to highlight the most influential features or data sources driving predictions. Compared to traditional machine learning, however, explainability in deep learning is still less mature and often provides only approximate insights into the decision process. For surgical applications, ensuring that these explanations are aligned with clinical reasoning remains crucial, as it allows clinicians to validate model predictions, detect potential biases, and build confidence in the safe integration of deep learning pipelines into perioperative practice.

## V. Conclusion

Despite the advancement of ML algorithms, this review highlights persistent methodological gaps in postoperative risk stratification and outcome prediction. Most studies rely on private, single-center datasets with limited modalities, restricting reproducibility, generalizability, and the integration of clinically available data such as time-series signals and clinical text. Reporting of key preprocessing steps, including imputation, feature selection, and class imbalance handling, was inconsistent and often incomplete. Conventional machine learning methods and shallow neural networks dominated, while few studies explored task-specific deep learning or multimodal integration. The lack of benchmark datasets and standardized evaluation protocols makes it difficult to compare models or track methodological progress. Explainability, essential for clinical trust and adoption, is not routinely addressed. These findings highlight the need for diverse and accessible datasets to support more robust surgical ML research. In parallel, future work should prioritize transparent reporting practices, methodological rigor, and clinically meaningful interpretability to improve the clinical utility and trustworthiness of predictive models. Future models may benefit from tailored architectures and the integration of multimodal clinical data to better capture perioperative complexity.

## Appendix

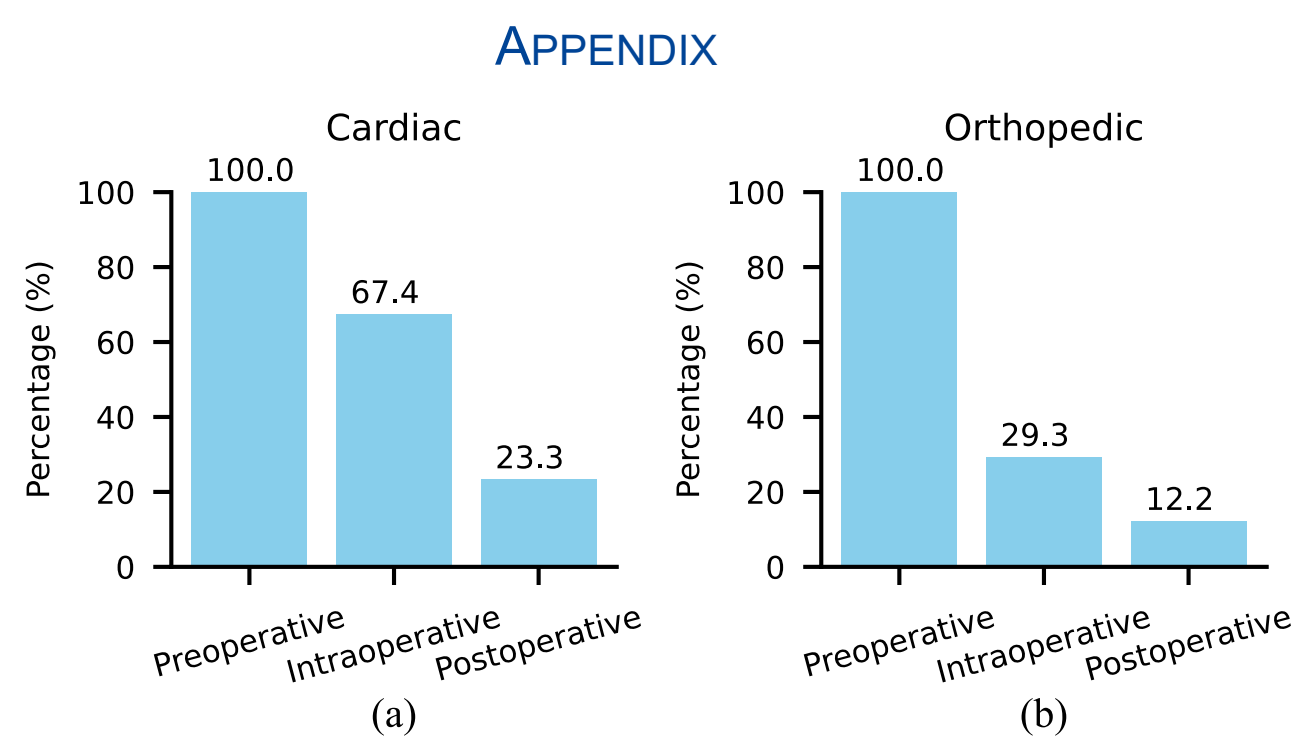

Fig. 5. Summary of input variables by data collection time frames in (a) cardiac, and (b) orthopedic surgeries.

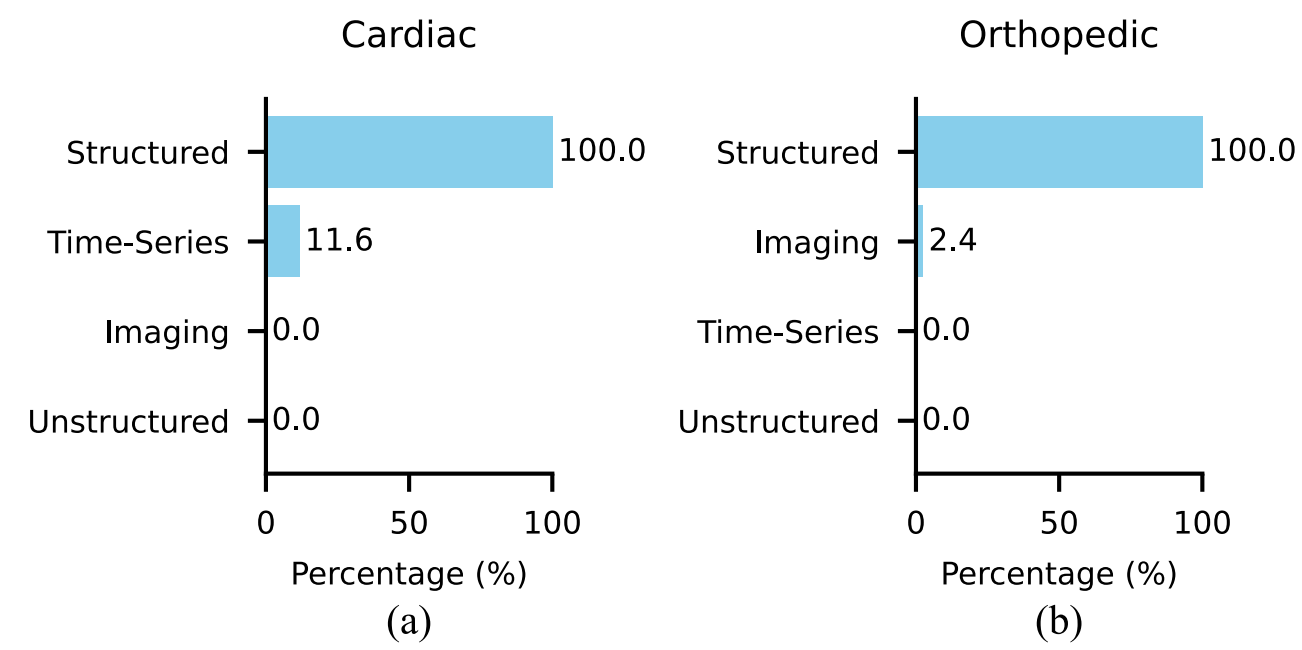

Fig. 6. Summary of input variables by data type in (a) cardiac, and (b) orthopedic surgeries.

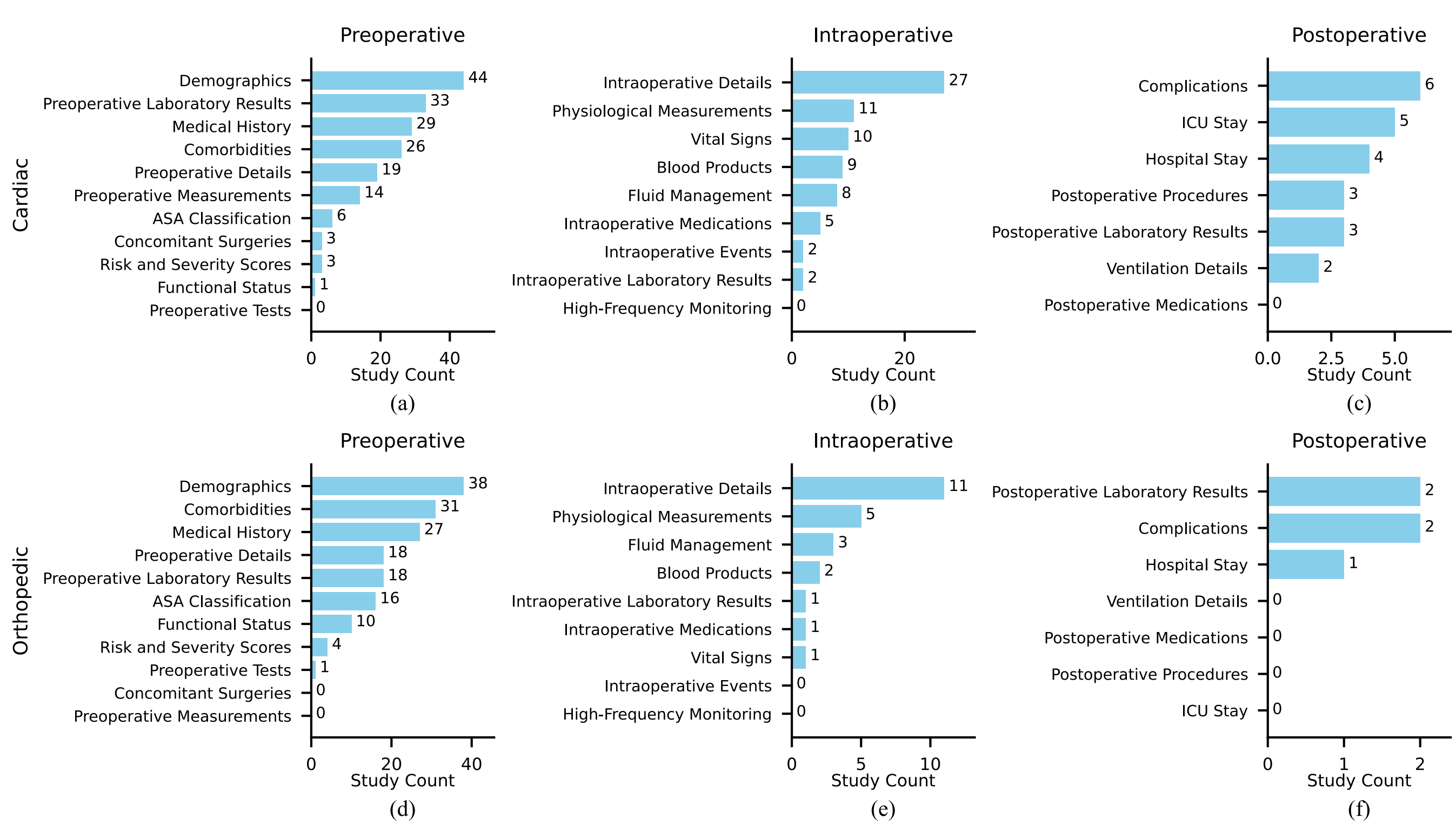

Fig. 7. Summary of input variables by clinical categories. The top row (a)–(c) represents cardiac surgery, and the bottom row (d)–(f) represents orthopedic surgery. Columns correspond to the preoperative (left), intraoperative (middle), and postoperative (right) periods, respectively.